\documentclass[11pt, a4paper, copyright, nonumbering]{mll_style}

\usepackage[numbers,sort]{natbib}
\setcitestyle{numbers,square,citesep={,}}
\usepackage{xspace}
\usepackage{adjustbox}
\usepackage{array}
\usepackage{float}
\usepackage{placeins}
\usepackage{dblfloatfix}
\usepackage{enumitem}
\usepackage{setspace}
\usepackage{multirow}
\usepackage{makecell}
\usepackage{longtable}
\usepackage{xltabular}
\usepackage{threeparttable}
\usepackage{siunitx}
\usepackage{pdflscape}
\usepackage{mathtools}
\usepackage{bm}
\usepackage{dsfont}
\usepackage{subcaption}
\usepackage{tikz}
\usepackage{nicematrix}
\usepackage{pgfplots}
\pgfplotsset{compat=1.18}
\usepackage{algorithm}
\usepackage{algpseudocode}
\usepackage{listings}
\usepackage[capitalize,noabbrev]{cleveref}
\usepackage{fontawesome5}
\usepackage[most]{tcolorbox}
\usepackage{CJKutf8}
\usepackage{pifont}
\usepackage{caption}
  \usepackage{etoolbox}
  \AtBeginEnvironment{tabular}{\scriptsize}

\usepackage[toc,page,header]{appendix}
\usepackage{minitoc}
\usepackage{titletoc}

\definecolor{blkA}{RGB}{245,247,255} %
\definecolor{blkB}{RGB}{245,250,245} %
\definecolor{blkC}{RGB}{255,247,245} %
\definecolor{gainGreen}{RGB}{24,130,72}
\definecolor{mutedGray}{RGB}{120,120,120}

\theoremstyle{plain}

\theoremstyle{definition}

\theoremstyle{remark}

\newif\ifUseAbbrev
\UseAbbrevfalse %

\reportnumber{001}

\fancypagestyle{msarfirst}[firststyle]{
    \fancyfoot[L,R]{}
    \fancyfoot[C]{%
        \parbox[t]{\textwidth}{\raggedright
            \normalfont\fontsize{8.5}{10}\selectfont
            $^\dagger$Corresponding author. Contact:
            \texttt{\{galaxy-1,yanjunchi\}@sjtu.edu.cn}\\[3pt]
            \makebox[\textwidth][c]{\footerfont\thepage}%
        }%
    }
}

\title{\centering Multi-Stream Attention Residuals\vspace{-8pt}}

\date{}

\author[*]{
~Xiaohan Qin$^{1,2,3}$,
Xiangdong Zhang$^{1,3}$,
Tuo Dai$^{3}$,
Sunan Zou$^{3,4}$,
Huaijin Wu$^{1,3}$,\authorcr
Yebin Yang$^{1,3}$,
Zhuo Xia$^{1,3}$,
Lin Yao$^{3}$,
Yuliang Liu$^{5}$,
Junchi Yan$^{1,\dagger}$\\
{\small $^1$School of AI, Shanghai Jiao Tong University~~~$^2$Shanghai Innovation Institute}\\
{\small $^3$Dots Studio, Xiaohongshu Inc.~~~$^4$School of CS, Peking University}\\
{\small $^5$Independent Researcher}\\
{\normalfont\small \textbf{Project page:}
\href{https://github.com/jianke0604/MSAR}{\nolinkurl{https://github.com/jianke0604/MSAR}}}\\
\vspace{-10pt}
}

\begin{abstract}
Attention Residuals (AttnRes) enables content-dependent retrieval over network
depth, which can be viewed as single-stream softmax attention along the depth axis.
Inspired by the success of multi-head attention along the sequence axis, we
introduce \textbf{Multi-Stream Attention Residuals (MSAR)}, an extension of
AttnRes that generalizes this mechanism to multi-head depth attention,
yielding substantial performance gains with minimal additional overhead.
MSAR draws stream diversity from sequence context through multi-scale causal
convolutions, allowing different heads to retrieve historical representations
through distinct contextual views while preserving the integrity of the hidden
representation.
The retrieved streams are combined through lightweight adaptive mixing to generate
the single-stream layer input.
By retaining single-stream sources and expanding them on demand, MSAR supports
multi-stream computation without requiring persistent multi-stream storage or
communication.
Across all 12 downstream benchmarks at the 18B and 28B MoE scales,
MSAR consistently outperforms prior residual variants, including standard
residual connections, manifold-constrained Hyper-Connections (mHC), and AttnRes.
Scaling-law experiments yield an iso-loss compute speedup of $\mathbf{1.58\times}$ over
the standard residual baseline, compared with $\mathbf{1.35\times}$ for AttnRes and
$\mathbf{1.30\times}$ for mHC.
These results make MSAR a promising residual architecture for large-scale
pre-training.
\end{abstract}

\begin{document}
\begin{CJK*}{UTF8}{gbsn}

\doparttoc
\faketableofcontents

    \maketitle
    \thispagestyle{msarfirst}

\section{Introduction}
\label{sec:intro}

Scaling depth has been a primary lever for improving Transformer-based large
language models (LLMs)~\citep{yang2025qwen3,team2025kimi,liu2025deepseek}.
In modern PreNorm architectures~\citep{xiong2020layer}, however,
adding layers does not always yield the expected gains~\citep{gromov2024unreasonable,men2025shortgpt}.
The residual stream~\citep{he2016deep} aggregates preceding computations through a
single hidden-state trajectory with fixed additive updates; as the stack grows
deeper, features formed in early layers can be progressively diluted and become
difficult for later layers to retrieve.
This depth-wise information dilution limits the effective use of additional
layers, making depth scaling less efficient than its representational capacity
would suggest~\citep{sun2025curse,csordas2025language,AttnRes}.
Motivated by this, recent work has begun to challenge the standard
residual connection design itself.
Representatively, Hyper-Connections (HC), together with its manifold-constrained
variant mHC, expands the single residual stream into multiple parallel streams with learned stream
mixing~\citep{zhu2024hyper,DS_mHC}, while Attention Residuals (AttnRes) replaces
fixed residual accumulation with learned depth-wise softmax retrieval over earlier
representations~\citep{AttnRes}. 
From the time-depth duality perspective of AttnRes~\citep{AttnRes,zhang2026residual}, these methods can be viewed as
attention-like mechanisms along the depth axis: HC-family methods maintain a mixed
multi-stream state, resembling recurrent, linear-attention-style depth memory,
whereas AttnRes instantiates single-head depth-wise softmax attention.

Inspired by the success of multi-head attention along the sequence axis~\citep{vaswani2017attention},
a natural question emerges: can we realize \emph{multi-head depth attention},
bringing greater expressiveness to residual retrieval over depth?
However, previous attempts indicate that achieving this goal is non-trivial.
AttnRes already tried channel-splitting the hidden state to form multi-head depth
aggregation, which instead degraded performance~\citep{AttnRes}: in depth-wise
attention the retrieved value is essentially the hidden state itself and should be
treated as a coherent whole, so slicing it into low-dimensional fragments breaks this
integrity. Another intuitive route is to derive several candidate streams from a single
layer output using the $\mathcal{H}^{\mathrm{post}}$-style write-back of (m)HC, and then
treat them as depth-attention heads.
This keeps each candidate full-dimensional, but the candidates are effectively
scalar reweightings of the same vector; normalization largely removes
these magnitude differences, yielding nearly identical keys and limiting
the diversity of source representations across heads.
Taken together, these observations point to a central design challenge:
how to construct diverse depth-attention heads from a single layer output
without compromising its integrity.

\begin{figure}[t]
    \centering
    \captionsetup[subfigure]{justification=centering,singlelinecheck=true}
    \makebox[\textwidth][c]{%
        \begin{subfigure}[t]{0.32\textwidth}
            \centering
            \includegraphics[width=\linewidth]{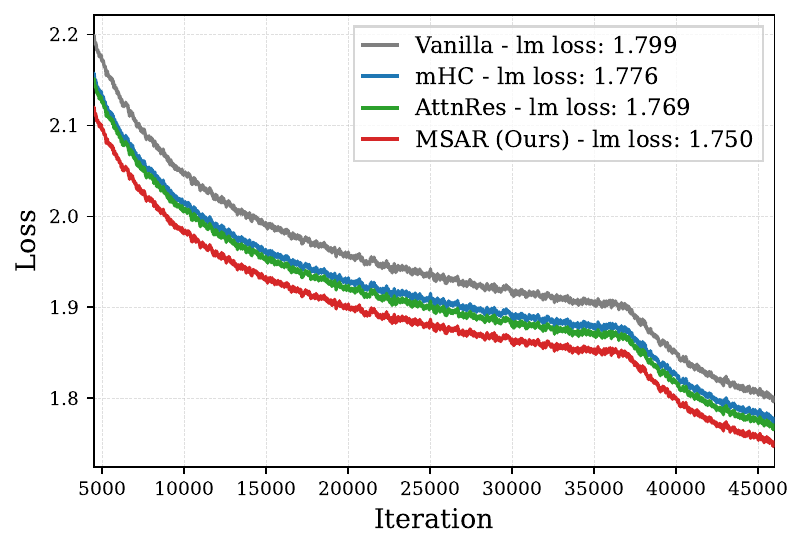}
            \caption{Training loss.}
            \label{fig:intro_loss}
        \end{subfigure}
        \hspace{0.01\textwidth}
        \begin{subfigure}[t]{0.32\textwidth}
            \centering
            \includegraphics[width=\linewidth]{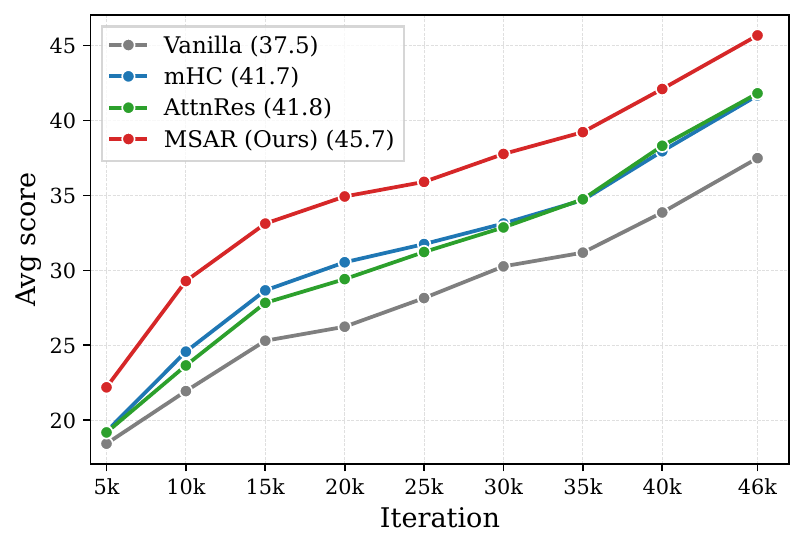}
            \caption{Average downstream score.}
            \label{fig:intro_avg_score}
        \end{subfigure}
        \hspace{0.01\textwidth}
        \begin{subfigure}[t]{0.32\textwidth}
            \centering
            \includegraphics[width=\linewidth]{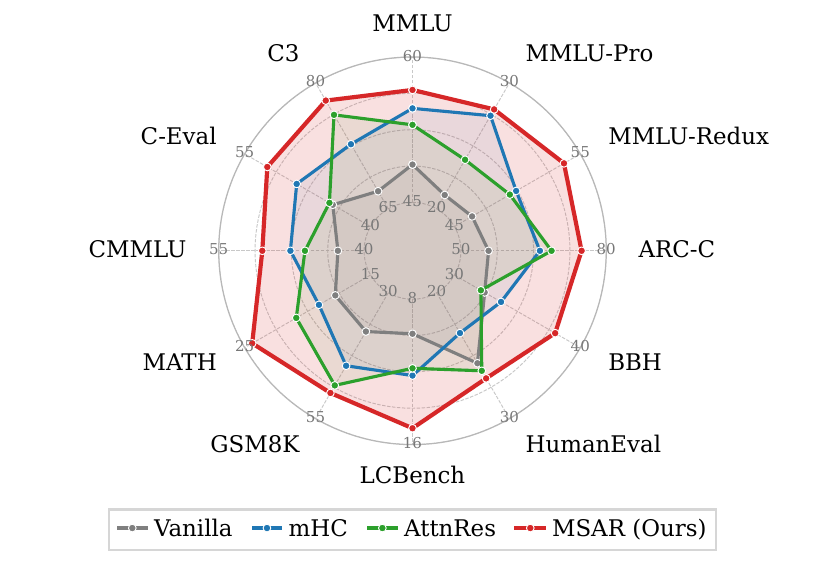}
            \caption{Per-benchmark performance.}
            \label{fig:intro_benchmark_scores}
        \end{subfigure}%
    }
    \caption{\textbf{MSAR improves optimization and downstream performance at
    18B scale.} MSAR achieves lower training loss and higher average downstream
    score than baseline residual variants.}
    \vspace{-10pt}
    \label{fig:intro_results}
\end{figure}

We therefore propose \textbf{M}ulti-\textbf{S}tream \textbf{A}ttention
\textbf{R}esiduals (\textbf{MSAR}), a low-cost extension of
AttnRes that couples contextual stream expansion with depth-wise retrieval
to address this challenge.
Starting from a single residual source, MSAR draws diversity from the sequence
axis: it applies a small set of multi-scale causal depthwise convolutions to the
layer output, producing $m$ full-dimensional streams whose receptive fields encode
different local-context scales.
These streams serve as distinct source views for depth attention, so each stream
can perform its own softmax retrieval over depth instead of sharing one depth
distribution.
The $m$ retrieved states are then fused by a lightweight
$\mathcal{H}^{\mathrm{pre}}$-style mixer into the input of the target layer.
In this way, MSAR turns residual retrieval into a multi-head depth-attention
primitive: different heads can select different depth histories through distinct
sequence-context views while preserving the integrity of the hidden representation.

At the system level, MSAR is fully compatible with AttnRes block mode,
cache-based pipeline communication, and the two-phase computation strategy.
In our 28B MoE setting, MSAR increases training FLOPs by only $0.5\%$ and
activated parameters by $0.3\%$.
Notably, during the forward pass, all streams are deterministically generated from the
main residual state through causal depthwise convolutions, allowing MSAR to keep
only the main stream in memory and expand it on demand when computing depth
attention.
The convolution-expanded source streams are intermediate values used only by
depth attention and are immediately consumed by the following attention
computation; in practice, this allows stream expansion and consumption to be
handled by a fused kernel, avoiding write-back of these $m$ intermediate streams.
The same principle applies in block mode: each block accumulates layer outputs
into a single-stream summary, and the $m$ streams are recomputed from these
summaries when they are consumed by depth attention.
With this design, for $N$ stored sources, MSAR reduces pipeline communication and
prefill activation storage from $O(mNd)$ to $O(Nd)$, matching the storage cost of
AttnRes.
Compared with AttnRes, the remaining overhead mainly comes from the memory
access needed to write the retrieved multi-stream outputs and aggregate these
streams.

Empirically, MSAR consistently improves both training loss and downstream
performance in large-scale pre-training.
As summarized in Figure~\ref{fig:intro_results}, in our 18B MoE setting, MSAR
achieves the lowest final training loss, reaching $1.750$ compared with Vanilla
($1.799$), mHC ($1.776$), and AttnRes ($1.769$).
This training-loss gain consistently translates into higher average downstream
scores throughout training across representative benchmarks covering knowledge,
reasoning, math, and code. 
Scaling-law experiments further confirm that MSAR reaches the same loss with
substantially less compute: relative to the Vanilla baseline, MSAR achieves an
iso-loss compute speedup of $1.58\times$, compared with
$1.35\times$ for AttnRes and $1.30\times$ for mHC.
These results suggest that enriching depth-wise residual retrieval with
multi-stream sequence-context views is an effective route for improving depth
scaling. Our contributions can be summarized as follows:

\textbf{1). Multi-Stream Attention Residuals.}
We introduce MSAR, which generalizes AttnRes from single-stream residual
retrieval to multi-head depth attention.
MSAR uses low-cost multi-scale causal depthwise convolutions to construct
multiple streams, performs per-stream depth attention, and aggregates the
retrieved streams with lightweight scalar mixing.
This yields stream diversity without breaking the full-dimensional coherence of
the hidden representation.

\textbf{2). Efficient design.}
MSAR stores only single-stream summaries and expands multi-stream representations
on demand, reducing pipeline communication and prefill activation storage from
$O(mNd)$ to $O(Nd)$ and matching the storage cost of AttnRes.
It remains fully compatible with AttnRes's cross-stage caching and two-phase
inference strategy, with only minimal training FLOPs and parameter overhead.

\textbf{3). Large-scale validation.}
We validate MSAR in large-scale pre-training, where it delivers significant and
consistent downstream gains in both 18B and 28B MoE settings.
Scaling-law experiments and ablation studies further verify the effectiveness of
the design, showing that the gains come from the proposed inductive bias rather
than from a model-size-specific artifact.

\section{Related Work}
\label{sec:related}

\textbf{Multi-State Residual Recurrence.}
One approach improves depth-wise information flow by maintaining multiple
residual states~\citep{zhang2026ddl,li2026siamesenorm,yang2026mhclite,sHC,zheng2026multi}.
Hyper-Connections (HC)~\citep{zhu2024hyper} implement this through learned
stream aggregation, output distribution, and inter-stream mixing.
Its manifold-constrained variant, mHC~\citep{DS_mHC},
stabilizes multi-stream recurrence through doubly stochastic residual mixing,
while its Single-Pass variant~\citep{deepseekai2026deepseekv41flash} removes
the dependency on current-state read coefficients to enable single-pass
fused execution.
xHC~\citep{zhang2026xhc} enriches the write-back signal
and uses sparse stream operations to support larger residual
expansion.
Gated Residual~\citep{qiu2026design} uses elementwise gated reads and
input-dependent scalar writes over multiple residual streams, without an
explicit inter-stream residual-mixing operator.

\textbf{Cross-Layer Access.}
Another line of work gives later layers explicit access to earlier
representations~\citep{zhang2026ancre,fang2023cross,menghani2025laurel,zhou2025valueresiduallearning,peters-etal-2018-deep}.
DenseNet~\citep{densenet} connects layers by concatenating preceding feature maps.
DenseFormer~\citep{pagliardini2024denseformer} aggregates earlier hidden
states using learned, input-independent depth weights.
MUDDFormer~\citep{xiao2025muddformer} introduces input-dependent, multiway
aggregation of historical layer representations.
Mixture-of-Depths Attention (MoDA)~\citep{MoDA} jointly attends to
sequence key-value pairs and same-position key-value pairs from preceding
layers.
Attention Residuals (AttnRes)~\citep{AttnRes} replaces residual accumulation
with softmax-weighted retrieval over earlier layer outputs, using block-level
sources and specialized execution strategies to reduce storage and
communication costs.
Representative methods from both lines, mHC and AttnRes, have also been
validated in trillion-parameter-scale pre-training through
DeepSeek-V4-Pro~\citep{xu2026deepseek} and
Kimi K3~\citep{team2026kimi}, respectively.

\textbf{Position of MSAR.}
MSAR instantiates the intersection of the two residual-design axes above.
It preserves the explicit cross-layer access of AttnRes, while introducing a
multi-stream retrieval structure inspired by the capacity expansion of
HC-family methods.
Rather than widening the persistent residual recurrence, MSAR widens the
depth-attention primitive itself by exposing historical hidden states through
multiple sequence-context views.

\section{Method}
\label{sec:method}

\subsection{Preliminaries: Residuals as Depth Mixing}
\label{sec:method-preliminaries}

\textbf{Notation.}
We index depth at the granularity of residual branches: each self-attention or
MLP branch in a Transformer stack is counted as one layer.
Unless otherwise stated, equations are written for a single sequence position,
with batch and token indices suppressed.
Let $\bm{h}_l \in \mathbb{R}^{d}$ denote the representation provided as input to
the $l$-th residual branch, for $l \in \{1,\ldots,L\}$, and let $\bm{h}_1$ be
the token embedding.
We write $f_l(\cdot)$ for the transformation performed by branch $l$, absorbing
the local normalization, attention or MLP computation, and output projection into
this operator.

\textbf{Standard Residual Accumulation.}
A standard residual stack propagates representations by adding each branch output
to the running hidden state:
\begin{equation}
    \bm{h}_{l+1}
    =
    \bm{h}_{l} + f_l(\bm{h}_l).
\end{equation}
We treat the embedding as the initial depth source and define
\begin{equation}
    \bm{v}_0 = \bm{h}_1,
    \qquad
    \bm{v}_i = f_i(\bm{h}_i), \quad i \geq 1.
\end{equation}
Unrolling the recurrence gives
\(
    \bm{h}_l
    =
    \bm{h}_1 + \sum_{i=1}^{l-1} f_i(\bm{h}_i)
    =
    \sum_{i=0}^{l-1}\bm{v}_i.
\)
Thus, standard residual connections perform depth mixing with fixed unit
coefficients: every previous source is accumulated into a single residual stream.
It preserves a direct identity path across layers, allowing information and
gradients to pass through the stack without being forced through every nonlinear
transformation.

The same formulation also prescribes how information is aggregated across depth.
All previous updates are compressed into one running state, so later layers read
the accumulated representation rather than individual depth sources.
Moreover, as more updates are added, the residual state grows with depth, making
each new update progressively smaller relative to the accumulated stream.

\textbf{(Manifold-Constrained) Hyper-Connections.}
A direct way to relax this single-state constraint is to expand the residual
state into multiple persistent streams.
This increases the capacity of the residual path to carry distinct depth
histories, yielding significant pre-training improvements with modest additional
FLOPs~\citep{zhu2024hyper,DS_mHC}.
Let
\begin{equation}
    \mathbf{X}_l
    =
    \left(\bm{x}_{l}^{(1)},\ldots,\bm{x}_{l}^{(m)}\right)^\top
    \in \mathbb{R}^{m \times d}
\end{equation}
denote the multi-stream state before layer $l$, where $m$ is the number of
streams.
Hyper-Connections first read a layer input from this state and then write the
layer output back to all streams:
\begin{gather}
    \bm{h}_l
    =
    \left(\mathcal{H}^{\mathrm{pre}}_l \mathbf{X}_l\right)^\top,
    \qquad
    \bm{v}_l
    =
    f_l(\bm{h}_l),
    \\
    \mathbf{X}_{l+1}
    =
    \mathcal{H}^{\mathrm{res}}_l \mathbf{X}_l
    +
    \mathcal{H}^{\mathrm{post}}_l f_l(\bm{h}_l)^\top .
\end{gather}
Here $\mathcal{H}^{\mathrm{pre}}_l \in \mathbb{R}^{1 \times m}$ aggregates
streams into the layer input, $\mathcal{H}^{\mathrm{post}}_l \in
\mathbb{R}^{m \times 1}$ distributes the new residual update across streams, and
$\mathcal{H}^{\mathrm{res}}_l \in \mathbb{R}^{m \times m}$ mixes the persistent
state~\citep{zhu2024hyper}.
The manifold-constrained variant mHC further projects
$\mathcal{H}^{\mathrm{res}}_l$ onto the Birkhoff polytope of doubly stochastic
matrices via the Sinkhorn--Knopp algorithm, preserving stable signal propagation
while retaining the multi-state residual form~\citep{DS_mHC}.

\textbf{Attention Residuals.}
A complementary route is to expose individual sources instead of only the
accumulated state.
For layer $l$, AttnRes replaces the residual sum with a normalized mixture over
the embedding source and all preceding branch outputs:
\begin{equation}
    \bm{h}_l
    =
    \alpha_{0 \to l}\bm{h}_1
    +
    \sum_{i=1}^{l-1}
    \alpha_{i \to l} f_i(\bm{h}_i).
\end{equation}
The coefficients $\alpha_{i \to l}$ are layer-specific attention weights
satisfying $\sum_{i=0}^{l-1}\alpha_{i \to l}=1$; they are obtained by
applying a softmax over all available sources:
\begin{equation}
    \alpha_{i \to l}
    =
    \frac{
    \phi(\bm{q}_l,\bm{k}_i)
    }{
    \sum_{j=0}^{l-1}\phi(\bm{q}_l,\bm{k}_j)
    } .
\end{equation}
AttnRes instantiates this attention with a layer-specific learned pseudo-query
and uses each residual source as its key:
\begin{equation}
    \bm{q}_l = \bm{w}_l,
    \qquad
    \bm{k}_i = \bm{v}_i,
    \qquad
    \phi(\bm{q},\bm{k})
    =
    \exp\left(
    \bm{q}^{\top}\operatorname{RMSNorm}(\bm{k})
    \right).
\end{equation}
This turns residual aggregation into content-dependent depth attention by
explicitly preserving previous hidden states as retrievable sources~\citep{AttnRes}.

\subsection{MSAR: Multi-Stream Attention Residuals}
\label{sec:mars}

\textbf{Multi-Stream Source Expansion.}
As discussed in Section~\ref{sec:intro}, multi-head depth attention requires
a non-trivial expansion of a single layer output into multiple streams with
sufficient head diversity.
A natural approach is to generate these streams through separate dense
projections, but this substantially increases the parameter count and FLOPs.
Instead, we borrow additional contextual information from neighboring tokens
to construct multiple views of each residual source at low cost.

Under autoregressive modeling with a causal mask, earlier hidden states in the
same layer encode the context available at their respective positions and can
therefore supplement the current source without exposing future information.
Aggregating these states over different local windows provides source views
with different contextual coverage while preserving the integrity of the hidden
representation. Specifically, we use lightweight causal depthwise 1D convolutions to expand each
residual source into $m$ streams:
\begin{equation}
    \bm{v}_i^{(r)}
    =
    \mathcal{E}_{i,r}[\bm{v}_i],
    \qquad r=0,\ldots,m-1,
    \label{eq:mars-source-expansion}
\end{equation}
where $\mathcal{E}_{i,0}=\mathrm{Id}$ and $\mathcal{E}_{i,r}$ for $r\geq 1$
is a layer-specific causal depthwise convolution with kernel size $\kappa_r$.
Restoring the token index $t$ gives
\begin{equation}
    \bm{v}_{i,t}^{(0)}=\bm{v}_{i,t},
    \qquad
    \bm{v}_{i,t}^{(r)}
    =
    \sum_{\delta=0}^{\kappa_r-1}
    \bm{c}_{i,r,\delta}\odot\bm{v}_{i,t-\delta},
    \quad r=1,\ldots,m-1,
    \label{eq:mars-causal-expansion}
\end{equation}
where $\bm{c}_{i,r,\delta}\in\mathbb{R}^{d}$ contains the channel-wise
convolution weights, and positions before the sequence start are zero-padded.
The kernel sizes assign different local receptive fields to the streams:
the original stream preserves the current source, while the additional streams
incorporate preceding representations over different ranges.
In our main setting, we retain the original source and apply three
convolutions with kernel sizes $\{4,8,12\}$, yielding $m=4$ streams.

\textbf{Per-Stream Depth Attention.}
The expanded streams serve as distinct depth-attention heads, each associated
with a different contextual scale.
For each target layer, we perform depth attention independently within each
stream, aggregating the corresponding source representations across depth:
\begin{equation}
    \bm{z}_l^{(r)}
    =
    \alpha_{0\to l}^{(r)}\mathcal{E}_{0,r}[\bm{h}_1]
    +
    \sum_{i=1}^{l-1}
    \alpha_{i\to l}^{(r)}\mathcal{E}_{i,r}[f_i(\bm{h}_i)]
    =
    \sum_{i=0}^{l-1}\alpha_{i\to l}^{(r)}\bm{v}_i^{(r)},
    \qquad r=0,\ldots,m-1.
    \label{eq:mars-depth-retrieval}
\end{equation}
Each target layer now has $m$ independently learned pseudo-queries, one per stream.
The attention weights are defined as
\begin{equation}
    \bm{q}_l^{(r)}=\bm{w}_l^{(r)},
    \qquad
    \alpha_{i\to l}^{(r)}
    =
    \frac{
    \phi(\bm{q}_l^{(r)},\bm{k}_i^{(r)})
    }{
    \sum_{j=0}^{l-1}\phi(\bm{q}_l^{(r)},\bm{k}_j^{(r)})
    },
    \label{eq:mars-depth-weights}
\end{equation}
where $\bm{k}_i^{(r)}=\bm{v}_i^{(r)}$, and $\phi$ is the same RMS-normalized exponential kernel defined
in Section~\ref{sec:method-preliminaries}.
The weights are normalized over historical sources within each stream,
so $\sum_{i=0}^{l-1}\alpha_{i\to l}^{(r)}=1$ for every $r$.
This produces $m$ retrieved states
$\mathbf{Z}_l=[\bm{z}_l^{(0)},\ldots,\bm{z}_l^{(m-1)}]^\top
\in\mathbb{R}^{m\times d}$.

\textbf{Stream Fusion.}
The retrieved streams $\mathbf{Z}_l$ must be fused into a single input for the
target layer.
Following the output projection of sequence-level multi-head attention would
require an $md\times d$ matrix.
Instead, we use a lightweight mHC-style $\mathcal{H}^{\mathrm{pre}}$
mixer~\citep{DS_mHC} to generate input-dependent scalar weights:
\begin{equation}
    \mathcal{H}^{\mathrm{pre}}_l
    =
    \sigma\!\left(
    \alpha_l^{\mathrm{pre}}
    \operatorname{RMSNorm}\!\left(\operatorname{Flatten}(\mathbf{Z}_l)\right)
    \mathbf{W}_l^{\mathrm{pre}}
    +\bm{b}_l^{\mathrm{pre}}
    \right)
    \in\mathbb{R}^{1\times m}.
    \label{eq:mars-pre-mixer}
\end{equation}
Here $\mathbf{W}_l^{\mathrm{pre}}\in\mathbb{R}^{md\times m}$ is a lightweight
projection, $\alpha_l^{\mathrm{pre}}$ is a learnable scaling factor,
$\bm{b}_l^{\mathrm{pre}}\in\mathbb{R}^{1\times m}$ is a per-stream bias,
and $\sigma$ denotes the sigmoid function.
The target layer input is a weighted sum of the retrieved streams:
\begin{equation}
    \bm{h}_l
    =
    \left(\mathcal{H}^{\mathrm{pre}}_l\mathbf{Z}_l\right)^\top
    =
    \sum_{r=0}^{m-1}
    \left[\mathcal{H}^{\mathrm{pre}}_l\right]_r
    \bm{z}_l^{(r)}
    \in\mathbb{R}^{d}.
    \label{eq:mars-stream-fusion}
\end{equation}
For the last layer output, we omit further expansion for depth retrieval,
directly perform a residual merge and pass the result to the LM head.

\begin{figure}[!t]
    \centering
    \includegraphics[width=\textwidth]{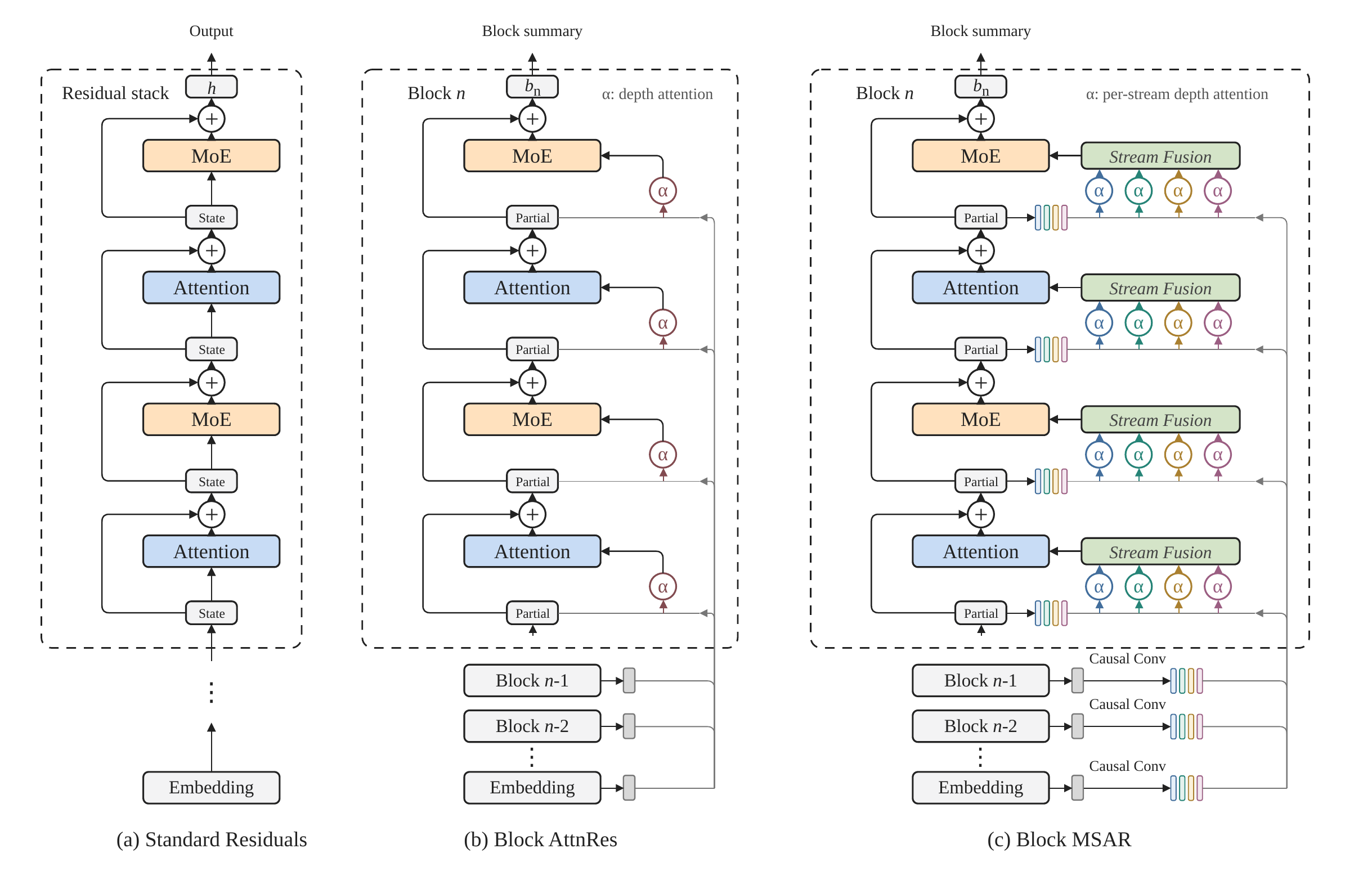}
    \caption{\textbf{Comparison of residual architectures.}
    (a) Standard Residuals; (b) Block AttnRes; (c) Block MSAR.
    Block MSAR expands single-stream sources into contextual streams for
    per-stream depth attention, then fuses the retrieved streams into a
    single input for each residual branch.}
    \label{fig:residual-architecture}
    \vspace{-10pt}
\end{figure}

\subsection{Block MSAR}
\label{sec:block-mars}

Directly applying MSAR requires retaining the hidden state produced by every
layer, incurring considerable memory and pipeline communication overhead in
large-scale training.
Following AttnRes~\citep{AttnRes}, we introduce Block MSAR, which replaces
individual layer outputs with block-level summaries as depth-attention sources.
We partition the $L$ residual branches into $N$ contiguous blocks
$\mathcal{B}_1,\ldots,\mathcal{B}_N$.
Within each block, outputs accumulate in a single stream; the resulting
summaries are then expanded for multi-stream depth attention.

We retain the embedding as the initial source $\bm{b}_0=\bm{h}_1$ and define
the completed summary of block $n$ as
\begin{equation}
    \bm{b}_n
    =
    \sum_{j\in\mathcal{B}_n} f_j(\bm{h}_j)
    \in\mathbb{R}^{d}.
    \label{eq:mars-block-summary}
\end{equation}
For a target layer $l\in\mathcal{B}_n$, the current block contributes the
partial sum of its preceding outputs,
\begin{equation}
    \bm{b}_n^{<l}
    =
    \sum_{\substack{j\in\mathcal{B}_n, j<l}} f_j(\bm{h}_j),
    \qquad
    \bm{b}_n^{<l+1}
    =
    \bm{b}_n^{<l}+f_l(\bm{h}_l).
    \label{eq:mars-block-partial}
\end{equation}
The partial sum starts at zero and becomes $\bm b_n$ when the block completes.
Stream expansion is applied to the accumulated single-stream representation.
Below, $\mathcal E_r$ omits bank indices.

For layer $l\in\mathcal{B}_n$, each head retrieves from the embedding,
completed block summaries, and the current partial after source expansion:
\begin{equation}
    \bm{z}_l^{(r)}
    =
    \sum_{j=0}^{n-1}
    \alpha_{j\to l}^{(r)}\mathcal{E}_r[\bm{b}_j]
    +
    \alpha_{n\to l}^{(r)}\mathcal{E}_r[\bm{b}_n^{<l}],
    \qquad r=0,\ldots,m-1.
    \label{eq:mars-block-retrieval}
\end{equation}
Here the source index in $\alpha_{j\to l}^{(r)}$ refers to a block summary,
with index $n$ denoting the current partial.
The layer-specific pseudo-queries and key construction follow
Section~\ref{sec:mars}, with softmax normalization over all available sources
within each stream.
At the first layer of a block, the partial source is absent and is omitted
from both the retrieval sum and the softmax denominator.
The retrieved states form $\mathbf{Z}_l$ and are fused into $\bm{h}_l$
using Eq.~\ref{eq:mars-stream-fusion}.
Figure~\ref{fig:residual-architecture} contrasts this design with
standard residual connections and Block AttnRes.

\subsection{Efficiency Analysis}
\label{sec:efficiency-analysis}

\textbf{Single-Stream Storage and Communication.}
Since the expanded streams are deterministic functions of the single-stream
block summaries defined in Section~\ref{sec:block-mars}, MSAR only needs to
retain and communicate single-stream sources.
These sources are expanded through lightweight causal convolutions only when
depth attention is computed.
To preserve this property in Block MSAR, we deliberately accumulate layer
outputs before stream expansion, rather than expand each output and
accumulate streams separately.
For $N$ summaries, this reduces persistent source storage and the corresponding
pipeline communication payload per token from $O(mNd)$ to $O(Nd)$, matching
Block AttnRes under the same pipeline schedule.
The single-stream representation also supports AttnRes's cross-stage
caching~\citep{AttnRes}, allowing previously received summaries to be reused
across virtual pipeline stages without retransmission.

\textbf{Two-Phase Computation.}
All target layers within a block access the same completed sources, and their
learned pseudo-queries are available before the block executes.
This enables a two-phase computation strategy for MSAR, analogous to
AttnRes, with each stream processed independently.
In the first phase, queries for all target layers in the block are batched
against the completed sources to compute per-target,
per-stream retrieval results and softmax normalization statistics.
Batching allows the targets to share source reads, reducing repeated memory
access across layers.
In the second phase, each layer computes the contribution of its current
partial and merges it with its precomputed result through online softmax.
This recovers the attention in Eq.~\ref{eq:mars-block-retrieval}; the first
layer of a block uses only the completed-source result.
Stream fusion is applied after the merge, so the first phase retains $m$
retrieved streams per target, separately from the single-stream source storage.

\textbf{Fused Stream Expansion.}
On-demand expansion can introduce additional memory traffic if its
outputs are written out and subsequently read by attention.
The expanded streams are consumed immediately by the subsequent depth attention
and have no other consumers in the forward pass.
This allows causal convolutional expansion and per-stream attention to be
combined in a fused kernel, avoiding writes of these intermediate source
streams to memory.
This complements two-phase computation: batching reduces repeated source reads
across targets, while fusion removes intermediate writes and reads during
stream expansion.
The final retrieved multi-stream outputs are still written and subsequently
read for $\mathcal{H}^{\mathrm{pre}}$ aggregation, contributing additional
memory traffic relative to AttnRes.
Appendix~\ref{app:cost-analysis} compares the per-token, per-layer memory
access costs of the residual mechanisms across methods.

\section{Experiments}
\label{sec:exp}

\subsection{Experimental Setup}
\label{sec:exp-setup}

\textbf{Training Settings.}
We conduct language model pre-training experiments with DeepSeekMoE-style
Transformers~\citep{dai2024deepseekmoe} using SwiGLU feed-forward
networks~\citep{shazeer2020glu}.
Our main results cover two MoE scales with 18B and 28B total parameters,
and we use a 10B MoE model for ablation studies.
Comprehensive training details are provided in
Appendix~\ref{app:training-details}.
Unless otherwise specified, we use Block MSAR with $m=4$ streams and
$N\approx 8$ blocks.
We compare MSAR with two representative residual architectures,
mHC~\citep{DS_mHC} with $m=4$ streams and AttnRes~\citep{AttnRes} with
$N\approx 8$ blocks.
We also report a Transformer with standard residual connections as the
Vanilla baseline.

\textbf{Evaluation.}
We evaluate downstream performance on 12 benchmarks grouped into three categories:
general capabilities (MMLU~\citep{MMLU}, MMLU-Pro~\citep{wang2024mmlupro},
MMLU-Redux~\citep{redux}, ARC-C~\citep{ARCC}, and BBH~\citep{BBH});
code and mathematics (HumanEval~\citep{humaneval},
LCBench~\citep{2023opencompass}, GSM8K~\citep{gsm8k}, and MATH~\citep{math});
and Chinese language understanding (CMMLU~\citep{li2024cmmlu},
C-Eval~\citep{ceval}, and C3~\citep{c3}).
Evaluation protocols and few-shot settings are provided in
Appendix~\ref{app:training-details}.

\subsection{Main Results}
\label{sec:exp-main-results}

We compare MSAR against mHC, AttnRes, and the Vanilla residual baseline on
18B and 28B MoE models.
Table~\ref{tab:main-results} summarizes downstream results under comparable
parameter counts and training FLOPs.
In our 28B MoE setting, MSAR with $m=4$ streams and $N\approx 8$ blocks adds
only $0.3\%$ activated parameters and $0.5\%$ training FLOPs
relative to Vanilla.
Detailed parameterization and overhead comparisons across methods are
provided in Appendix~\ref{app:cost-analysis}.

\begin{table*}[t]
    \centering
    \small
    \caption{Downstream performance comparison of MSAR with Vanilla, mHC,
    and AttnRes on 18B and 28B MoE models.
    Scores are reported in percent (higher is better), with the best result
    for each benchmark at each scale in \textbf{bold}.}
    \label{tab:main-results}
    \begin{tabular}{l cccc cccc}
        \toprule
        & \multicolumn{4}{c}{\textbf{18B}}
        & \multicolumn{4}{c}{\textbf{28B}} \\
        \cmidrule(lr){2-5} \cmidrule(lr){6-9}
        Benchmark & Vanilla & mHC & AttnRes & MSAR
        & Vanilla & mHC & AttnRes & MSAR \\
        \midrule
        \multicolumn{9}{l}{\textit{General}} \\
        \quad MMLU       & 48.9 & 54.7 & 53.0 & \textbf{56.6} & 54.6 & 56.8 & 59.9 & \textbf{62.5} \\
        \quad MMLU-Pro   & 21.1 & 27.4 & 23.9 & \textbf{27.9} & 30.1 & 34.9 & 36.0 & \textbf{38.7} \\
        \quad MMLU-Redux & 46.4 & 49.9 & 49.4 & \textbf{53.7} & 50.6 & 53.9 & 55.7 & \textbf{58.7} \\
        \quad ARC-C      & 55.7 & 66.3 & 68.7 & \textbf{74.9} & 70.8 & 74.9 & 75.6 & \textbf{78.0} \\
        \quad BBH        & 32.4 & 33.7 & 32.1 & \textbf{38.0} & 41.7 & 43.6 & 40.7 & \textbf{46.1} \\
        \midrule
        \multicolumn{9}{l}{\textit{Code \& Math}} \\
        \quad HumanEval  & 25.6 & 23.2 & 26.2 & \textbf{26.8} & 27.4 & 26.8 & 29.3 & \textbf{31.1} \\
        \quad LCBench    &  9.9 & 12.2 & 11.8 & \textbf{15.1} & 15.1 & 14.8 & 16.5 & \textbf{18.4} \\
        \quad GSM8K      & 37.7 & 44.5 & 48.4 & \textbf{49.9} & 50.3 & 56.3 & 57.2 & \textbf{65.7} \\
        \quad MATH       & 17.8 & 19.1 & 20.9 & \textbf{24.4} & 21.9 & 26.5 & 28.0 & \textbf{28.8} \\
        \midrule
        \multicolumn{9}{l}{\textit{Chinese}} \\
        \quad CMMLU      & 42.7 & 47.6 & 46.1 & \textbf{50.5} & 47.6 & 50.1 & 51.3 & \textbf{56.4} \\
        \quad C-Eval     & 44.5 & 48.8 & 44.9 & \textbf{52.3} & 50.2 & 51.2 & 59.1 & \textbf{59.9} \\
        \quad C3         & 67.1 & 72.7 & 76.2 & \textbf{77.9} & 75.2 & 78.7 & 83.5 & \textbf{84.9} \\
        \midrule
        \textbf{Average} & 37.5 & 41.7 & 41.8 & \textbf{45.7} & 44.6 & 47.4 & 49.4 & \textbf{52.4} \\
        \bottomrule
    \end{tabular}
\end{table*}

As shown in Table~\ref{tab:main-results}, MSAR outperforms mHC, AttnRes,
and the Vanilla baseline on all 12 evaluated benchmarks at both model scales.
In the 18B setting, MSAR achieves an average score of 45.7, compared with
37.5 for Vanilla (+8.2), 41.7 for mHC (+4.0), and 41.8 for AttnRes (+3.9).
While both mHC and AttnRes improve substantially over standard residual
connections, MSAR provides a further gain over both approaches.
Representative gains over AttnRes include MMLU (+3.5),
ARC-C (+6.2), and BBH (+5.9).
The gains persist in the larger 28B setting, where MSAR achieves an average
score of 52.4, compared with 44.6 for Vanilla (+7.8), 47.4 for mHC (+5.0),
and 49.4 for AttnRes (+3.0).
This consistent downstream advantage across both model scales further
supports the effectiveness of MSAR in large-scale pre-training.

\suppressfloats[t]
\begin{figure}[t]
    \centering
    \begin{minipage}[t]{0.58\textwidth}
        \vspace{0pt}
        \centering
        \includegraphics[width=\linewidth]{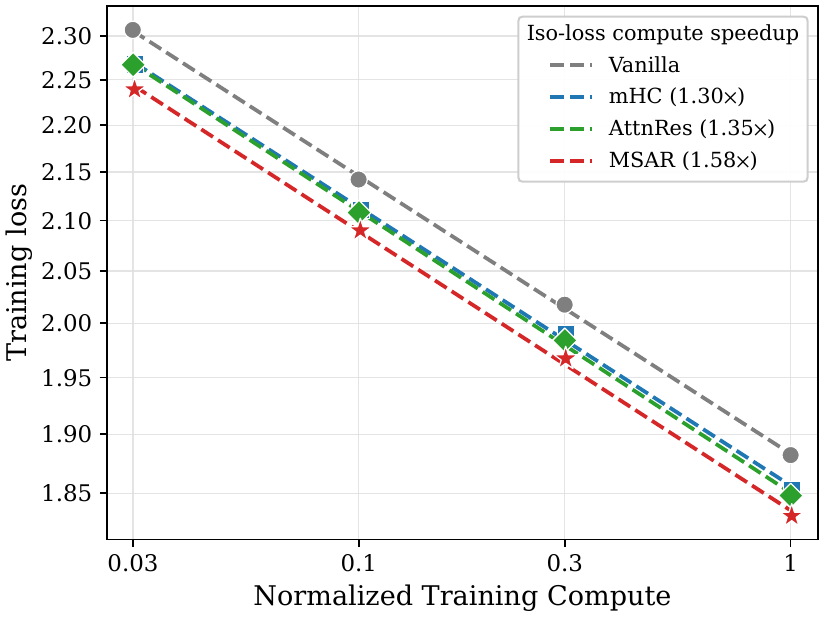}
        \caption{Scaling-law comparison of MSAR, AttnRes, mHC, and Vanilla.
        MSAR substantially outperforms all three baselines in the
        loss--compute trade-off.}
        \label{fig:scaling-laws}
    \end{minipage}\hfill
    \begin{minipage}[t]{0.4\textwidth}
        \vspace{0pt}
        \centering
        \captionof{table}{Ablations on a 10B MoE model.
        $m$: number of streams; $S$: residual sublayers per block.}
        \label{tab:ablations}
        \footnotesize
        \setlength{\tabcolsep}{3pt}
        \renewcommand{\arraystretch}{1.12}
        \begin{tabularx}{\linewidth}{@{}>{\raggedright\arraybackslash}Xccr@{}}
            \toprule
            Variant & $m$ & $S$ & Pile loss $\downarrow$ \\
            \midrule
            (1) Vanilla          & -- & -- & 1.898 \\
            (2) mHC              & 4  & -- & 1.871 \\
            (3) AttnRes          & 1  & 4  & 1.867 \\
            (4) AttnRes + mHC    & 4  & 4  & 1.861 \\
            \midrule
            \rowcolor{black!6}
            (5) \textbf{MSAR (default)} & 4 & 4 & 1.847 \\
            (6) w/o Conv   & 4  & 4  & 1.862 \\
            (7) w/o $\mathcal{H}^{\mathrm{pre}}$
                                 & 4  & 4  & 1.851 \\
            \midrule
            (8) $m=2$             & 2  & 4  & 1.850 \\
            (9) $m=8$             & 8  & 4  & 1.846 \\
            \midrule
            (10) Full MSAR       & 4  & 1  & 1.843 \\
            (11) $S=2$           & 4  & 2  & 1.845 \\
            (12) $S=8$           & 4  & 8  & 1.852 \\
            \bottomrule
        \end{tabularx}
    \end{minipage}
\end{figure}

\subsection{Scaling Laws}
\label{sec:exp-scaling-laws}

To quantify the compute efficiency of MSAR across model scales, we compare
its training-loss scaling with Vanilla, mHC, and AttnRes.
We evaluate four model scales, matching training-token budgets across methods
at each scale.
Our FLOPs accounting excludes the LM head and includes method-specific
residual overhead.
Training compute is normalized by the Vanilla training FLOPs at the largest
evaluated model scale, using the same reference for all methods.
We fit a shifted power law to the final training loss:
\begin{equation}
    \mathcal{L}(C) = A C^{-\beta} + E,
    \label{eq:scaling-law}
\end{equation}
where $C$ denotes normalized training compute, $A$ and $\beta$ are fitted separately for
each method, and $E=0.72$ is the estimated irreducible loss, which represents
the asymptotic loss floor as training compute increases.
We estimate the parameters by linear regression in log space.
Detailed model configurations are provided in
Appendix~\ref{app:scaling-law-details}.

Figure~\ref{fig:scaling-laws} shows that MSAR achieves the lowest training
loss at all four scales, and its fitted curve remains below all three
baselines over their shared compute range.
At the largest model scale, MSAR reduces training loss by $2.7\%$ relative
to Vanilla, compared with reductions of $1.6\%$ for mHC and $1.8\%$ for
AttnRes.
Averaged across the four model scales, MSAR achieves an iso-loss compute
speedup of $1.58\times$ relative to Vanilla, compared with $1.35\times$ for
AttnRes and $1.30\times$ for mHC.
These results support a consistent improvement in the loss--compute
trade-off across the evaluated scales.

\subsection{Ablation Studies}
\label{sec:exp-ablations}

We evaluate the key design choices of MSAR on a 10B MoE model with
15 Transformer layers, comprising 30 residual sublayers.
Table~\ref{tab:ablations} reports Pile loss~\citep{gao2020pile}, with $S$ denoting the number
of residual sublayers per block.
Detailed model configurations are provided in
Appendix~\ref{app:training-details}.

\textbf{Comparison with Prior Methods.}
Table~\ref{tab:ablations}, rows (1)--(5), compares MSAR with standard
residual connections and prior residual architectures.
mHC and AttnRes reduce loss from the Vanilla baseline's 1.898 to 1.871 and
1.867, respectively, while MSAR achieves a substantially lower loss of 1.847.
We also evaluate a hybrid residual architecture, AttnRes + mHC, that
simultaneously maintains AttnRes's historical sources for cross-layer
access and mHC's multi-stream recurrent state, summing their readouts to
form the layer input.
This direct combination reaches 1.861, an additional reduction of only
0.006 over AttnRes.
With the same $m=4$ and $S=4$, MSAR lowers loss by a further 0.014.
These results suggest that MSAR's advantage stems from its specific
design, rather than simply adding the states
produced by cross-layer access and multi-stream recurrence.

\textbf{Component Design.}
Rows (5)--(7) isolate the contributions of contextual source expansion
and adaptive stream fusion.
Removing the convolutions replicates each source into four identical
streams while retaining four independently learned pseudo-queries per
target layer, yielding a multi-query depth-attention variant over shared
source representations.
The independent queries can still produce distinct depth-attention
distributions, allowing this ablation to distinguish the benefit of
contextual source expansion from that of multiple retrieval heads alone.
This increases loss from 1.847 to 1.862, supporting the importance of
source diversity beyond query diversity alone.
Replacing $\mathcal{H}^{\mathrm{pre}}$ with mean aggregation produces a
relatively small loss degradation to 1.851, indicating that adaptive fusion provides
an additional benefit while uniform averaging retains much of the gain.
This modest contribution is consistent with the HC component ablations
reported in mHC~\citep{DS_mHC}, and our experiments also indicate that
$\mathcal{H}^{\mathrm{pre}}$
contributes to training stability.

\textbf{Hyperparameter Sensitivity.}
Rows (5) and (8)--(12) examine the number of streams and block granularity.
Two streams already capture most of the improvement over Vanilla,
achieving a loss of 1.850.
Increasing $m$ to four lowers loss to 1.847, while eight streams yield
only a marginal further improvement to 1.846.
We therefore use $m=4$ by default.
For block granularity, $S=2,4,8$ correspond to $N=15,8,4$ blocks,
respectively, excluding the embedding source.
Their losses are 1.845, 1.847, and 1.852, compared with 1.843 for Full MSAR.
Thus, retaining approximately eight blocks recovers most of the benefit
of full depth access, with diminishing gains from finer partitioning.
Consistent with the block-size ablations in AttnRes,
we use $N\approx 8$ as the default.

\section{Conclusion and Outlook}
\label{sec:conclusion}

We introduce Multi-Stream Attention Residuals (MSAR), which extends
single-stream residual retrieval to multi-head depth attention.
MSAR draws source diversity from the sequence axis through lightweight
multi-scale causal convolutions, retrieves independently over depth for
each stream, and aggregates the retrieved states with adaptive scalar mixing.
By storing single-stream summaries and expanding them on demand, Block MSAR
matches Block AttnRes's persistent source storage and pipeline communication
costs while supporting cross-stage caching and two-phase computation.
Experiments on 18B and 28B MoE models show consistent downstream improvements
over prior residual variants, while scaling-law experiments yield an
iso-loss compute speedup of $1.58\times$ relative to the standard residual
baseline.
These results establish MSAR as an effective residual
architecture for improving the quality and compute efficiency of language
model pre-training.
More broadly, our results suggest that the benefits of multi-state
representations and cross-layer access are not fully overlapping,
motivating more elegant residual designs that combine their complementary
strengths with efficient execution across diverse model architectures
and large-scale pre-training settings.

\clearpage

{
\small
\bibliographystyle{plainnat}
\bibliography{references}
}

\clearpage
\appendix

\section{Experimental Details}
\label{app:training-details}

\textbf{Model Architectures and Training Settings.}
We use a 10B MoE model for ablation studies and 18B and 28B MoE models for
the main comparisons.
All backbones follow the DeepSeekMoE-style architecture~\citep{dai2024deepseekmoe},
with one leading dense Transformer layer followed by MoE layers, grouped-query
attention, and SwiGLU feed-forward networks.
Table~\ref{tab:training-configurations} summarizes the model configurations
and training hyperparameters; layer counts refer to Transformer layers,
each containing an attention and an MLP residual sublayer.

\begin{table}[htbp]
    \centering
    \small
    \caption{Backbone configurations and training hyperparameters.
    The 10B model is used for ablation studies, and the 18B and 28B models
    are used for the main results.
    Activated parameter counts are approximate and exclude input embeddings.}
    \label{tab:training-configurations}
    \setlength{\tabcolsep}{5pt}
    \renewcommand{\arraystretch}{1.08}
    \begin{tabular}{lccc}
        \toprule
        \textbf{Attribute} & \textbf{10B} & \textbf{18B} & \textbf{28B} \\
        \midrule
        Total Parameters            & 10B   & 18B   & 28B \\
        Activated Parameters        & 1.1B  & 1.7B  & 2.7B \\
        \midrule
        Transformer Layers          & 15    & 28    & 32 \\
        Leading Dense Layers        & 1     & 1     & 1 \\
        MoE Layers                  & 14    & 27    & 31 \\
        Hidden Dimension            & 2080  & 2112  & 2560 \\
        FFN Dimension (dense layer) & 6144  & 6144  & 6144 \\
        Attention Heads             & 16    & 16    & 20 \\
        KV Groups (GQA)             & 8     & 8     & 10 \\
        Head Dimension              & 128   & 128   & 128 \\
        Routed Experts              & 144   & 144   & 144 \\
        Shared Experts              & 1     & 1     & 1 \\
        Active Experts (Top-$k$)    & 8     & 8     & 8 \\
        Expert FFN Dimension        & 704   & 672   & 768 \\
        Shared Expert Dimension     & 704   & 672   & 768 \\
        Normalization               & \multicolumn{3}{c}{RMSNorm ($\epsilon=10^{-5}$)} \\
        Position Embedding          & \multicolumn{3}{c}{RoPE ($\theta=50000$)} \\
        Activation                  & \multicolumn{3}{c}{SwiGLU} \\
        \midrule
        Sequence Length             & \multicolumn{3}{c}{8192} \\
        Vocabulary Size             & \multicolumn{3}{c}{152064} \\
        Tokenizer                   & \multicolumn{3}{c}{Qwen2} \\
        Global Batch Size           & 288   & -     & 768 \\
        Training Tokens             & -     & -     & - \\
        Warmup Steps                & 500   & 500   & 500 \\
        Optimizer                   & \multicolumn{3}{c}{AdamW ($\beta_1=0.9$, $\beta_2=0.95$, $\epsilon=10^{-15}$)} \\
        Base Learning Rate          & $4.82\times10^{-4}$ & $3.97\times10^{-4}$ & $3.5\times10^{-4}$ \\
        LR Schedule                 & \multicolumn{3}{c}{Warmup--stable--decay (exponential decay)} \\
        Min LR Ratio                & \multicolumn{3}{c}{0.1} \\
        Weight Decay                & \multicolumn{3}{c}{0.1} \\
        Gradient Clipping           & \multicolumn{3}{c}{1.0} \\
        \bottomrule
    \end{tabular}
\end{table}

\textbf{Residual Configurations.}
mHC uses $m=4$ residual streams, 20 Sinkhorn--Knopp iterations, and gating
scales initialized to $0.01$.
For the 18B and 28B main experiments, both AttnRes and MSAR group four
Transformer layers into each block, corresponding to $S=8$ residual sublayers
and yielding $N=7$ and $N=8$ blocks, respectively.
The embedding is included as an additional, separate source and is not counted
in $N$.
Unless otherwise specified, MSAR uses $m=4$ streams and initializes its
gating scale $\alpha_l^{\mathrm{pre}}$ to $0.01$, as in mHC.
The three causal depthwise convolution branches have kernel sizes
$\{4,8,12\}$, with convolution weights initialized to zero.
The learned pseudo-queries in both MSAR and AttnRes are also initialized
to zero.

\textbf{Evaluation Protocols.}
We use a lightweight evaluation suite based on OpenCompass~\citep{2023opencompass}.
Likelihood-based evaluation selects the candidate answer with the lowest
conditional perplexity for MMLU~\citep{MMLU} (5-shot),
MMLU-Redux~\citep{redux} (5-shot), CMMLU~\citep{li2024cmmlu} (5-shot),
C-Eval~\citep{ceval} (5-shot), ARC-C~\citep{ARCC} (25-shot),
and C3~\citep{c3} (3-shot).
Generation-based evaluation is used for MMLU-Pro~\citep{wang2024mmlupro}
(5-shot), BBH~\citep{BBH} (3-shot), GSM8K~\citep{gsm8k} (4-shot),
MATH~\citep{math} (4-shot), HumanEval~\citep{humaneval} (0-shot),
and LCBench~\citep{2023opencompass} (5-shot), with task-specific answer
scoring for reasoning and mathematics and execution-based pass@1 for code.

\section{Scaling-Law Experiment Details}
\label{app:scaling-law-details}

We evaluate Vanilla, mHC, AttnRes, and MSAR at four model scales with matched
backbones and training-token budgets at each scale.
Table~\ref{tab:scaling-configurations} lists the configurations used in
Figure~\ref{fig:scaling-laws}.
All models use one leading dense Transformer layer, 144 routed experts with
top-8 routing, one shared expert, and a sequence length of 8192;
common optimizer settings follow Appendix~\ref{app:training-details}.
mHC and MSAR use $m=4$ streams, while AttnRes and MSAR share the block sizes
listed in the table, yielding 8, 8, 8, and 7 blocks across the four scales.
The last block may be shorter, and the embedding is an additional source
excluded from these block counts.

\begin{table}[htbp]
    \centering
    \small
    \caption{Scaling-law configurations.
    All methods use matched backbones and training-token budgets at each scale.
    The block size is shared by AttnRes and MSAR.}
    \label{tab:scaling-configurations}
    \setlength{\tabcolsep}{3pt}
    \renewcommand{\arraystretch}{1.15}
    \resizebox{\textwidth}{!}{%
    \begin{tabular}{ccccccccccc}
        \toprule
        Total & Activated & \multirow{2}{*}{Layers}
        & Hidden & Dense & Expert & Q/KV & Training & Global & Learning
        & Layers per Block \\
        Parameters$^\dagger$ & Parameters$^\dagger$ &
        & Dimension & FFN & FFN & Heads & Budget & Batch Size & Rate
        & (AttnRes/MSAR) \\
        \midrule
        2.21B  & 0.34B & 15 & 1024 & 3104 & 320 & 8/4  & - & 160 & $6.95\times10^{-4}$ & 2 \\
        3.75B  & 0.46B & 23 & 1040 & 3072 & 352 & 8/4  & - & 208 & $5.74\times10^{-4}$ & 3 \\
        9.47B  & 1.10B & 15 & 2080 & 6144 & 704 & 16/8 & - & 288 & $4.82\times10^{-4}$ & 2 \\
        17.40B & 1.77B & 28 & 2112 & 6144 & 672 & 16/8 & - & 384 & $3.97\times10^{-4}$ & 4 \\
        \bottomrule
    \end{tabular}%
    }
    \par\vspace{3pt}
    \begin{minipage}{\textwidth}
        \footnotesize
        $^\dagger$Backbone parameter counts exclude input embeddings.
    \end{minipage}
\end{table}

We fit Eq.~\ref{eq:scaling-law} independently for each method using its final
training losses and corresponding normalized compute budgets across the four model scales.
With $E=0.72$ fixed for all methods, we estimate $A$ and $\beta$ by ordinary
least squares regression of $\log_2(\mathcal{L}-E)$ on $\log_{10}C$.
The iso-loss compute speedup is the ratio of the fitted Vanilla compute at
each observed loss to the corresponding method's normalized training compute, averaged
across the four model scales.

\section{Parameter, Compute, and Memory Costs}
\label{app:cost-analysis}

\textbf{Parameter Overhead.}
\label{app:parameter-overhead}
We count additional parameters relative to the standard residual backbone,
including biases and learned scales.
The residual modules use RMS normalization without separate learnable
affine parameters.
Here $L$ counts residual sublayers
(twice the number of Transformer layers), and $d$ is the hidden dimension.
For mHC, each sublayer has a projection with $md(m^2+2m)$ parameters,
$m^2+2m$ biases, and three scalar scales. AttnRes adds one $d$-dimensional
pseudo-query per sublayer and one for the final output read:
\begin{equation}
    \Delta P_{\mathrm{mHC}}=L\bigl[md(m^2+2m)+m^2+2m+3\bigr],
    \qquad
    \Delta P_{\mathrm{AttnRes}}=(L+1)d.
\end{equation}

MSAR adds $md$ query parameters and $m^2d+m+1$ mixer parameters per
sublayer.
Each convolution bank contributes $d\sum_{r=1}^{m-1}\kappa_r$ parameters.
Our configuration uses independent banks for the first $L-1$ sublayer
outputs and one for the embedding source; the last output is merged
without expansion.
The embedding uses $\mathcal E_{0,r}$.
In Block MSAR, completed summary $\bm b_j$ uses $\mathcal E_{e_j,r}$,
where $e_j=\max\mathcal B_j$ is the last sublayer of block $j$;
a nonempty partial $\bm b_n^{<l}$ uses $\mathcal E_{l-1,r}$.
Completed-source banks remain fixed across subsequent target layers.
Thus,
\begin{equation}
    \Delta P_{\mathrm{MSAR}}
    =L\left[\left(m^2+m+\sum_{r=1}^{m-1}\kappa_r\right)d+m+1\right].
\end{equation}
With $m=4$ and kernel sizes $\{4,8,12\}$, this becomes $L(44d+5)$,
compared with $L(96d+27)$ for mHC.
The resulting parameter increments at the 18B and 28B scales are
$5.204$M and $7.209$M for MSAR, $11.356$M and $15.730$M for mHC,
and $0.120$M and $0.166$M for AttnRes, respectively.

\textbf{Training FLOPs.}
\label{app:training-flops}
We count a multiply--add as two FLOPs and estimate training FLOPs as three
times the forward cost, excluding activation recomputation and optimizer
updates. For the main experiments, the backbone cost includes attention
and active FFN projections, causal attention, and the LM-head projection.
For residual overhead, we account for stream-mapping projections and
mixing in mHC using the training implementation's estimator.
For AttnRes, we include key normalization, query--key scoring, softmax,
value aggregation, and block accumulation.
For MSAR, the overhead comprises convolutional stream expansion,
per-stream depth attention, and lightweight stream fusion.
Table~\ref{tab:residual-costs} summarizes the estimated parameter and
training FLOPs overheads relative to the Vanilla backbone.

\begin{table}[htbp]
    \centering
    \small
    \caption{Residual-module parameter and training FLOPs overheads.
    $\Delta P/P_{\mathrm{act}}$ is relative to Vanilla's activated parameters
    (approximately 1.7B and 2.7B at the two scales);
    $\Delta F/F_{\mathrm{Vanilla}}$ is relative to its total training FLOPs,
    including the LM head.}
    \label{tab:residual-costs}
    \begin{tabular}{lrrrr}
        \toprule
        & \multicolumn{2}{c}{\textbf{18B}} & \multicolumn{2}{c}{\textbf{28B}} \\
        \cmidrule(lr){2-3} \cmidrule(lr){4-5}
        Method & $\Delta P/P_{\mathrm{act}}$ & $\Delta F/F_{\mathrm{Vanilla}}$
        & $\Delta P/P_{\mathrm{act}}$ & $\Delta F/F_{\mathrm{Vanilla}}$ \\
        \midrule
        mHC     & $0.67\%$ & $0.66\%$ & $0.58\%$ & $0.60\%$ \\
        AttnRes & $0.01\%$ & $0.10\%$ & $0.01\%$ & $0.10\%$ \\
        MSAR    & $0.31\%$ & $0.51\%$ & $0.27\%$ & $0.48\%$ \\
        \bottomrule
    \end{tabular}
\end{table}

\textbf{Storage, Communication, and Memory Access.}
\label{app:memory-costs}
As discussed in Section~\ref{sec:efficiency-analysis}, Block MSAR matches
Block AttnRes in persistent source storage and pipeline communication
under the same block partition and pipeline schedule.
Table~\ref{tab:residual-memory-access} compares activation I/O costs,
using optimized execution with kernel fusion for Block MSAR.
The costs for the other methods are taken from AttnRes~\citep{AttnRes}.
For Block MSAR, source expansion is fused with depth attention:
the expanded source streams are immediately consumed by online softmax
without being materialized in memory, and only the resulting $m$
retrieved streams are written back.
$\mathcal{H}^{\mathrm{pre}}$ gate computation is fused with the online merge.
The retrieved streams are then read for final aggregation, giving a
leading-order cost of $(N/S+4m+5)d$ per token per residual sublayer.

\begin{table}[htbp]
    \centering
    \small
    \setlength{\tabcolsep}{4pt}
    \renewcommand{\arraystretch}{1.15}
    \caption{Memory access cost per token per residual sublayer incurred
    by each residual mechanism during the forward pass.
    We report leading terms proportional to $d$, excluding parameter
    traffic and the internal I/O of $f_l$.
    Phase~1 costs are averaged across depth and amortized over the $S$
    targets in each block. The Typical column uses $N=8$ blocks,
    $S=16$ residual sublayers per block (128 residual sublayers in total,
    corresponding to 64 Transformer layers), and $m=4$ streams.}
    \label{tab:residual-memory-access}
    \resizebox{\textwidth}{!}{%
    \begin{tabular}{@{}llcccc@{}}
        \toprule
        \multirow{2}{*}{Method} & \multirow{2}{*}{Operation}
        & \multirow{2}{*}{Read} & \multirow{2}{*}{Write}
        & \multicolumn{2}{c}{Total I/O} \\
        \cmidrule(lr){5-6}
        & & & & Symbolic & Typical \\
        \midrule
        Vanilla & Residual merge & $2d$ & $d$ & $3d$ & $3d$ \\
        \midrule
        \multirow{5}{*}{mHC}
        & Compute mixing coefficients & $md$ & $-$
        & \multirow{5}{*}{$(8m+2)d$} & \multirow{5}{*}{$34d$} \\
        & Apply $\mathcal{H}^{\mathrm{pre}}$ & $md$ & $d$ & & \\
        & Apply $\mathcal{H}^{\mathrm{post}}$ & $d$ & $md$ & & \\
        & Apply $\mathcal{H}^{\mathrm{res}}$ & $md$ & $md$ & & \\
        & Residual merge & $2md$ & $md$ & & \\
        \midrule
        \multirow{2}{*}{Block AttnRes}
        & Phase 1 (amortized) & $\frac{N}{S}d$ & $d$
        & \multirow{2}{*}{$(N/S+5)d$} & \multirow{2}{*}{$5.5d$} \\
        & Phase 2 & $3d$ & $d$ & & \\
        \midrule
        \multirow{5}{*}{Block MSAR}
        & Phase 1: logits (amortized) & $\frac{N}{2S}d$ & $-$
        & \multirow{5}{*}{$(N/S+4m+5)d$} & \multirow{5}{*}{$21.5d$} \\
        & Phase 1: values (amortized) & $\frac{N}{2S}d$ & $md$ & & \\
        & Phase 2: partial update and scoring & $2d$ & $d$ & & \\
        & Phase 2: merge and $\mathcal{H}^{\mathrm{pre}}$ gate & $(m+1)d$ & $md$ & & \\
        & Stream aggregation & $md$ & $d$ & & \\
        \bottomrule
    \end{tabular}%
    }
\end{table}

\section{Learned Depth-Attention Patterns}
\label{app:attention-patterns}

\begin{figure}[htbp]
    \centering
    \includegraphics[width=\textwidth]{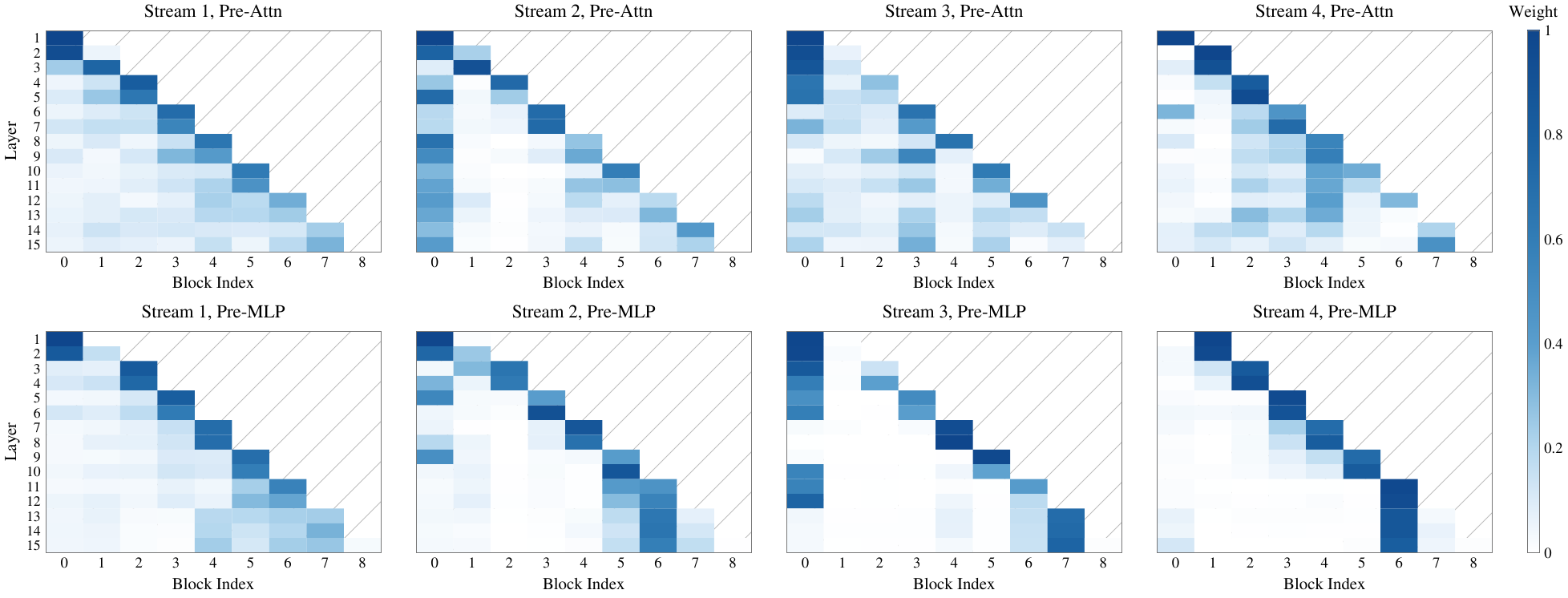}
    \caption{Token-averaged depth-attention weights of the 10B Block MSAR
    model. Rows show Pre-Attn (top) and Pre-MLP (bottom); columns show the
    four streams, numbered 1--4. Source 0 is the embedding, and sources 1--8 are blocks,
    with the current block represented by its nonempty partial sum.
    Hatched regions mark unavailable sources.}
    \label{fig:msar-attention-patterns}
\end{figure}

We visualize the learned depth-attention weights of the 10B Block MSAR
model in Figure~\ref{fig:msar-attention-patterns}.
The model has 15 Transformer layers, four streams, and eight blocks,
with the embedding retained as an additional source.
We average the pre-fusion attention weights over tokens from several
2,048-token Pile sequences and show Pre-Attn and Pre-MLP reads separately.

\textbf{Sublayer-Specific Access Modes.}
Pre-Attn reads distribute weight more broadly across depth, retaining
substantial access to the embedding and earlier blocks.
Pre-MLP reads instead tend to concentrate on fewer sources, producing
sharper near-diagonal and vertical structures.
The near-diagonal band corresponds to the current partial or most recent
completed block, while vertical bands indicate repeated access to a
particular source across target layers.

\textbf{Diverse Patterns across Streams.}
Within the same sublayer type, different streams develop distinct source
preferences.
In Pre-MLP reads beyond the first block, prominent attention to the
embedding is concentrated in Stream 2 (Conv-4) and Stream 3 (Conv-8), while
Stream 1 (identity) and Stream 4 (Conv-12) assign relatively little weight
to this source.
The streams also differ in which blocks they repeatedly access:
Stream 2 (Conv-4) and Stream 4 (Conv-12) exhibit strong vertical bands at
block 6, whereas Stream 3 (Conv-8) shows a pronounced preference for block 7.
These differences produce distinct vertical attention patterns across
streams, showing that MSAR learns diverse source-selection distributions
rather than repeating a single depth mixture.

\section{Discussion of Residual Mixing}
\label{app:residual-mixing}

\subsection{Comparison of Residual Mechanisms}
\label{app:residual-comparison}
\label{app:related-methods}

\begin{table}[!t]
    \centering
    \small
    \setlength{\tabcolsep}{5pt}
    \renewcommand{\arraystretch}{1.5}
    \caption{Comparison of residual update mechanisms.
    Weight distinguishes architecture-fixed, learned-static
    (input-independent after training), and input-dependent (dynamic)
    mixing coefficients.
    Source indicates the representations directly accessible to the
    residual mechanism. Branch normalization is absorbed into $f_l$;
    depth-attention weights are normalized over available sources.}
    \label{tab:residual-mechanisms}
    \resizebox{\textwidth}{!}{%
    \begin{tabular}{@{}llcl@{}}
        \toprule
        \textbf{Method} & \textbf{Update rule} & \textbf{Weight} & \textbf{Source} \\
        \midrule
        \multicolumn{4}{@{}l}{\emph{Single-state recurrence}} \\
        \addlinespace[2pt]
        Vanilla
        & $\bm h_l=\bm h_{l-1}+f_{l-1}(\bm h_{l-1})$
        & Fixed & $\bm h_{l-1}$ \\
        ReZero~\citep{bachlechner2020rezero}
        & $\bm h_l=\bm h_{l-1}+\alpha_l f_{l-1}(\bm h_{l-1})$
        & Static & $\bm h_{l-1}$ \\
        Highway~\citep{srivastava2015highway}
        & $\bm h_l=(1-\bm g_l)\odot\bm h_{l-1}+\bm g_l\odot f_{l-1}(\bm h_{l-1})$
        & Dynamic & $\bm h_{l-1}$ \\
        \midrule
        \multicolumn{4}{@{}l}{\emph{Multi-state recurrence}} \\
        \addlinespace[2pt]
        SiameseNorm~\citep{li2026siamesenorm}
        & $\bm h_l^1=\operatorname{Norm}(\bm h_{l-1}^1+\bm y_{l-1});\quad
          \bm h_l^2=\bm h_{l-1}^2+\bm y_{l-1}$
        & Fixed & $[\bm h_{l-1}^1,\bm h_{l-1}^2]$ \\
        HC/mHC~\citep{zhu2024hyper,DS_mHC}
        & $\mathbf H_l=\mathbf H_{l-1}\mathbf A_l
          +f_{l-1}(\mathbf H_{l-1}\bm\alpha_{l-1})\bm\beta_{l-1}^\top$
        & Dynamic & $\mathbf H_{l-1}$ \\
        DDL~\citep{zhang2026ddl}
        & $\mathbf H_l=(\mathbf I-\beta_l\bm k_l\bm k_l^\top)\mathbf H_{l-1}
          +\beta_l\bm k_l\bm v_l^\top$
        & Dynamic & $\mathbf H_{l-1}$ \\
        \midrule
        \multicolumn{4}{@{}l}{\emph{Cross-layer access}} \\
        \addlinespace[2pt]
        DenseNet~\citep{densenet}
        & $\bm h_l=\operatorname{ConvPool}([\bm h_1;f_1(\bm h_1);\ldots;f_{l-1}(\bm h_{l-1})])$
        & Static & $[\bm h_1,\ldots,\bm h_{l-1}]$ \\
        DenseFormer~\citep{pagliardini2024denseformer}
        & $\bm h_l=\alpha_{0\to l}\bm h_1+\sum_{i=1}^{l-1}\alpha_{i\to l}f_i(\bm h_i)$
        & Static & $[\bm h_1,\ldots,\bm h_{l-1}]$ \\
        Block AttnRes~\citep{AttnRes}
        & $\bm h_l=\sum_{i=0}^{n-1}\alpha_{i\to l}\bm b_i+\alpha_{n\to l}\bm b_n^j$
        & Dynamic & $[\bm b_0,\ldots,\bm b_{n-1},\bm b_n^j]$ \\
        \midrule
        \multicolumn{4}{@{}l}{\emph{Multi-stream cross-layer access}} \\
        \addlinespace[2pt]
        DAR~\citep{yu2026dualattentionresiduals}
        & $\bm h_l=g_l^{(0)}\sum_{i\in\mathcal I_l}\alpha_{i\to l}^{(0)}\bm c_i^0
          +g_l^{(1)}\sum_{i\in\mathcal I_l}\alpha_{i\to l}^{(1)}\bm c_i^1$
        & Dynamic & $[\mathbf H_0,\ldots,\mathbf H_{n-1},\mathbf P_n^j]$ \\
        Block MSAR
        & $\bm h_l=\sum_{r=0}^{m-1}g_l^{(r)}
          \sum_{\bm b\in\mathcal S_l}\alpha_l^{(r)}(\bm b)\mathcal E_r[\bm b]$
        & Dynamic & $\{\mathcal E_r[\bm b_0,\ldots,\bm b_{n-1},\bm b_n^j]\}_{r=0}^{m-1}$ \\
        \bottomrule
    \end{tabular}%
    }
\end{table}

Table~\ref{tab:residual-mechanisms} compares representative residual
mechanisms by their update rules, mixing weights, and accessible sources.
The formulations of several prior methods are adapted from
AttnRes~\citep{AttnRes}.

We use $\bm v_0$ for the embedding and $\bm v_i=f_i(\bm h_i)$ for
branch outputs.
ReZero uses a learned scalar $\alpha_l$, initialized to zero;
Highway uses an input-dependent gate $\bm g_l$; and DenseFormer uses
learned, input-independent coefficients $\alpha_{i\to l}$.
For DenseNet, $\operatorname{ConvPool}$ schematically denotes the
convolutional readout of concatenated features.
SiameseNorm adds the same branch output $\bm y_{l-1}$ to two states,
with normalization applied to one state.
In the DDL row, $\mathbf H_l$ denotes its matrix-valued state,
$\bm k_l$ and $\bm v_l$ denote its update key and value, and
$\beta_l$ controls the delta update; these symbols are local to that row.
For the block variants, layer $l$ belongs to block $n$
and accesses completed summaries together with the current partial.
The partial term is omitted at the first sublayer of a block.
We write $\bm b_n^j=\bm b_n^{<l}$ for the partial after the $j$
preceding sublayers of the current block.
For compactness, let
$\mathcal S_l=[\bm b_0,\ldots,\bm b_{n-1},\bm b_n^j]$
denote these available sources, with the empty partial excluded.
For MSAR, $\mathcal E_r[\mathcal S_l]$ denotes the list obtained by
expanding each source in stream $r$,
$\alpha_l^{(r)}(\bm b)$ is its normalized depth-attention weight, and
$g_l^{(r)}=[\mathcal H_l^{\mathrm{pre}}]_r$ is the MSAR fusion weight.
For HC/mHC, the column-stream notation is
$\mathbf H_l=\mathbf X_l^\top\in\mathbb R^{d\times m}$,
$\bm\alpha_l=(\mathcal H_l^{\mathrm{pre}})^\top$,
$\bm\beta_l=\mathcal H_l^{\mathrm{post}}$, and
$\mathbf A_l=(\mathcal H_{l-1}^{\mathrm{res}})^\top$.
Here $\bm\alpha_l$ is a stream-read vector, distinct from the scalar
coefficients used by ReZero, DenseFormer, and AttnRes.
DAR~\citep{yu2026dualattentionresiduals} similarly explores stream-width expansion for depth retrieval.
For DAR,
$\mathbf H_0,\ldots,\mathbf H_{n-1}$ are completed two-stream history
states and $\mathbf P_n^j$ is the current partial, omitted when empty.
In this row, each available candidate is written as
$\mathbf C_i=[\bm c_i^0,\bm c_i^1]$, with indices $i\in\mathcal I_l$.
The cross-stream weights satisfy
$\alpha_{i\to l}^{(0)}\propto\phi(\bm q_l^0,\bm c_i^1)$ and
$\alpha_{i\to l}^{(1)}\propto\phi(\bm q_l^1,\bm c_i^0)$,
each normalized over $\mathcal I_l$;
$\bm q_l^0,\bm q_l^1$ are learned pseudo-queries, and the sigmoid gates
$g_l^{(0)},g_l^{(1)}$ combine the two retrieved states.
The DAR row summarizes this retrieval; its stored states are updated
through gated branch writes and block-local stream mixing.
Its two streams maintain distinct residual trajectories at the same token position.
MSAR instead constructs retrieval sources from neighboring positions
through multi-scale causal convolutions, introducing head diversity
through distinct sequence-context views.
Each head can therefore select depth histories using a different
contextual scale.

\subsection{Residual Connections as Structured Matrices}
\label{app:structured-mars}

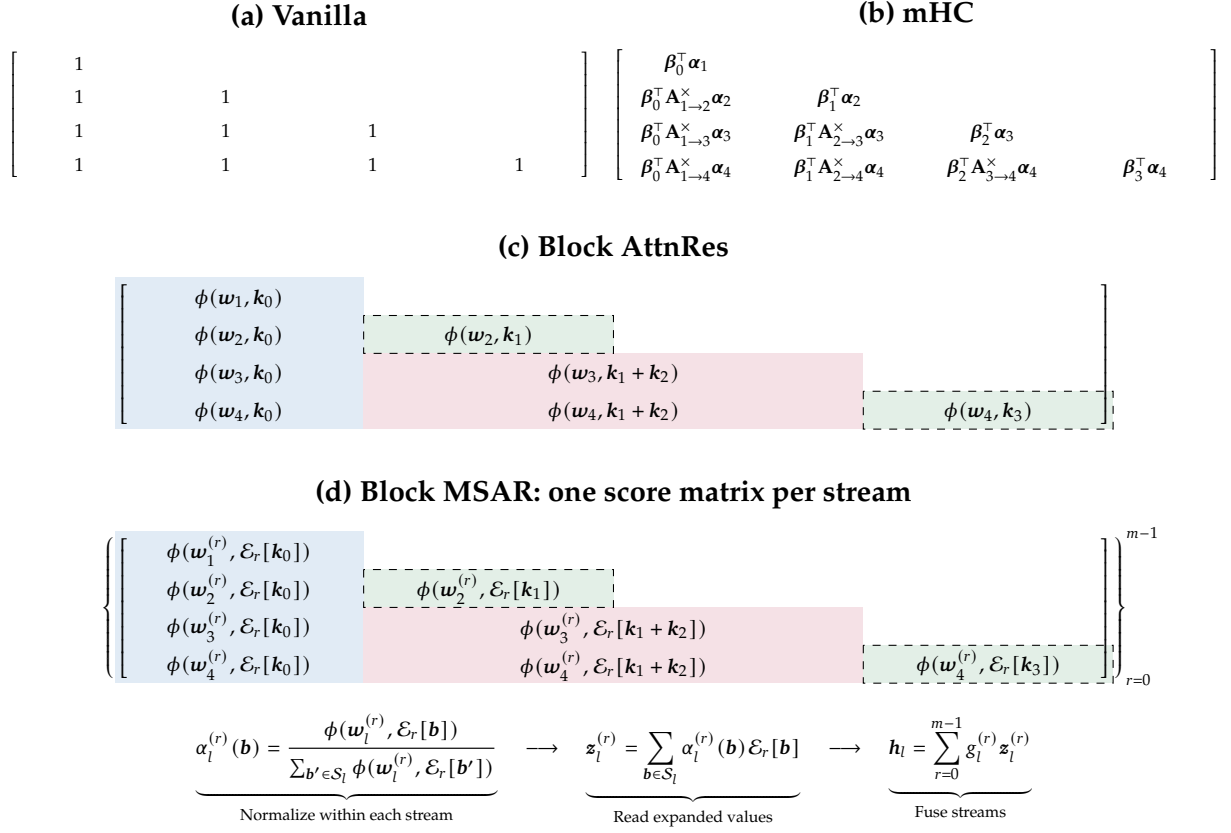
\begin{figure}[htbp]
    \centering
    \definecolor{MixEmbedding}{HTML}{E1EBF5}
    \definecolor{MixCompleted}{HTML}{F5E1E6}
    \definecolor{MixPartial}{HTML}{E3EFE7}
    \newcommand{\mixcell}[1]{\makebox[6.5em][c]{$#1$}}
    \newcommand{\marsmixcell}[1]{\makebox[10.5em][c]{$#1$}}
    \newsavebox{\marsdepthmatrix}
    \renewcommand{\arraystretch}{1.5}

    \begin{minipage}{0.48\textwidth}
        \centering
        \textbf{(a) Vanilla}\\[6pt]
        \scriptsize
        \resizebox{\linewidth}{!}{$\begin{bNiceArray}{cccc}
            \mixcell{1} & \mixcell{} & \mixcell{} & \mixcell{}\\
            \mixcell{1} & \mixcell{1} & \mixcell{} & \mixcell{}\\
            \mixcell{1} & \mixcell{1} & \mixcell{1} & \mixcell{}\\
            \mixcell{1} & \mixcell{1} & \mixcell{1} & \mixcell{1}
        \end{bNiceArray}$}
    \end{minipage}\hfill
    \begin{minipage}{0.50\textwidth}
        \centering
        \textbf{(b) mHC}\\[6pt]
        \scriptsize
        \resizebox{\linewidth}{!}{$\begin{bNiceArray}{cccc}
            \mixcell{\bm\beta_0^\top\bm\alpha_1}
            & \mixcell{} & \mixcell{} & \mixcell{}\\
            \mixcell{\bm\beta_0^\top\mathbf A_{1\to2}^{\times}\bm\alpha_2}
            & \mixcell{\bm\beta_1^\top\bm\alpha_2} & \mixcell{} & \mixcell{}\\
            \mixcell{\bm\beta_0^\top\mathbf A_{1\to3}^{\times}\bm\alpha_3}
            & \mixcell{\bm\beta_1^\top\mathbf A_{2\to3}^{\times}\bm\alpha_3}
            & \mixcell{\bm\beta_2^\top\bm\alpha_3} & \mixcell{}\\
            \mixcell{\bm\beta_0^\top\mathbf A_{1\to4}^{\times}\bm\alpha_4}
            & \mixcell{\bm\beta_1^\top\mathbf A_{2\to4}^{\times}\bm\alpha_4}
            & \mixcell{\bm\beta_2^\top\mathbf A_{3\to4}^{\times}\bm\alpha_4}
            & \mixcell{\bm\beta_3^\top\bm\alpha_4}
        \end{bNiceArray}$}
    \end{minipage}

    \vspace{18pt}

    \begin{minipage}{\textwidth}
        \centering
        \textbf{(c) Block AttnRes}\\[6pt]
        \scriptsize
        $\begin{bNiceArray}{cccc}
            \CodeBefore
            \begin{tikzpicture}
                \fill[MixEmbedding] (1-|1) rectangle (5-|2);
                \fill[MixPartial] (2-|2) rectangle (3-|3);
                \fill[MixCompleted] (3-|2) rectangle (5-|4);
                \fill[MixPartial] (4-|4) rectangle (5-|5);
                \draw[dashed, line width=0.4pt] (2-|2) rectangle (3-|3);
                \draw[dashed, line width=0.4pt] (4-|4) rectangle (5-|5);
            \end{tikzpicture}
            \Body
            \marsmixcell{\phi(\bm w_1,\bm k_0)}
            & \marsmixcell{} & \marsmixcell{} & \marsmixcell{}\\
            \marsmixcell{\phi(\bm w_2,\bm k_0)}
            & \marsmixcell{\phi(\bm w_2,\bm k_1)}
            & \marsmixcell{} & \marsmixcell{}\\
            \marsmixcell{\phi(\bm w_3,\bm k_0)}
            & \Block{1-2}{\phi(\bm w_3,\bm k_1+\bm k_2)}
            & & \marsmixcell{}\\
            \marsmixcell{\phi(\bm w_4,\bm k_0)}
            & \Block{1-2}{\phi(\bm w_4,\bm k_1+\bm k_2)}
            & & \marsmixcell{\phi(\bm w_4,\bm k_3)}
        \end{bNiceArray}$
    \end{minipage}

    \vspace{18pt}

    \begin{minipage}{\textwidth}
        \centering
        \textbf{(d) Block MSAR: one score matrix per stream}\\[6pt]
        \scriptsize
        \sbox{\marsdepthmatrix}{$\begin{bNiceArray}{cccc}
            \CodeBefore
            \begin{tikzpicture}
                \fill[MixEmbedding] (1-|1) rectangle (5-|2);
                \fill[MixPartial] (2-|2) rectangle (3-|3);
                \fill[MixCompleted] (3-|2) rectangle (5-|4);
                \fill[MixPartial] (4-|4) rectangle (5-|5);
                \draw[dashed, line width=0.4pt] (2-|2) rectangle (3-|3);
                \draw[dashed, line width=0.4pt] (4-|4) rectangle (5-|5);
            \end{tikzpicture}
            \Body
            \marsmixcell{\phi(\bm w_1^{(r)},\mathcal E_r[\bm k_0])}
            & \marsmixcell{} & \marsmixcell{} & \marsmixcell{}\\
            \marsmixcell{\phi(\bm w_2^{(r)},\mathcal E_r[\bm k_0])}
            & \marsmixcell{\phi(\bm w_2^{(r)},\mathcal E_r[\bm k_1])}
            & \marsmixcell{} & \marsmixcell{}\\
            \marsmixcell{\phi(\bm w_3^{(r)},\mathcal E_r[\bm k_0])}
            & \Block{1-2}{\phi(\bm w_3^{(r)},\mathcal E_r[\bm k_1+\bm k_2])}
            & & \marsmixcell{}\\
            \marsmixcell{\phi(\bm w_4^{(r)},\mathcal E_r[\bm k_0])}
            & \Block{1-2}{\phi(\bm w_4^{(r)},\mathcal E_r[\bm k_1+\bm k_2])}
            & & \marsmixcell{\phi(\bm w_4^{(r)},\mathcal E_r[\bm k_3])}
        \end{bNiceArray}$}%
        $\mathllap{\left\{\vphantom{\usebox{\marsdepthmatrix}}\right.}%
        \usebox{\marsdepthmatrix}%
        \mathrlap{\left.\vphantom{\usebox{\marsdepthmatrix}}\right\}_{r=0}^{m-1}}$
        \\[10pt]
        $\displaystyle
        \underbrace{\alpha_l^{(r)}(\bm b)
        =\frac{\phi(\bm w_l^{(r)},\mathcal E_r[\bm b])}
        {\sum_{\bm b'\in\mathcal S_l}\phi(\bm w_l^{(r)},\mathcal E_r[\bm b'])}}
        _{\text{Normalize within each stream}}
        \quad\longrightarrow\quad
        \underbrace{\bm z_l^{(r)}
        =\sum_{\bm b\in\mathcal S_l}\alpha_l^{(r)}(\bm b)\mathcal E_r[\bm b]}
        _{\text{Read expanded values}}
        \quad\longrightarrow\quad
        \underbrace{\bm h_l=\sum_{r=0}^{m-1}g_l^{(r)}\bm z_l^{(r)}}
        _{\text{Fuse streams}}$
    \end{minipage}

    \caption{Depth mixing for four residual mechanisms with
    $L=4$ and block size $S=2$. Blank entries are zero.
    Panels (a)--(b) show effective mixing coefficients; panels
    (c)--(d) show unnormalized $\phi$ scores.
    Blue denotes the embedding source, pink a completed block, and
    dashed green cells a current partial.
    A score spanning two columns corresponds to one block source.
    Block MSAR expands each source before scoring and value aggregation,
    normalizes separately within each stream, and fuses the retrieved states.}
    \label{fig:residual-depth-matrices}
\end{figure}

Following AttnRes~\citep{AttnRes}, Figure~\ref{fig:residual-depth-matrices}
visualizes depth mixing with $L=4$ residual sublayers and block size $S=2$.
Rows correspond to target inputs $\bm h_1,\ldots,\bm h_4$ and columns
to sources $\bm v_0,\ldots,\bm v_3$.
Vanilla assigns unit weights to every preceding source.
For mHC, unrolling the recurrence gives
\begin{equation}
    M_{i\to l}
    =\bm\beta_i^\top\mathbf A_{i+1\to l}^{\times}\bm\alpha_l,
    \qquad
    \mathbf A_{a\to b}^{\times}
    =\mathbf A_{a+1}\mathbf A_{a+2}\cdots\mathbf A_b.
    \label{eq:discussion-mhc-matrix-entry}
\end{equation}
The product is ordered in increasing depth, with an empty product
equal to the identity. For the embedding, $\bm\beta_0=\bm 1_m$
and the transition product starts at $\mathbf A_2$.
Each entry therefore contains the source write vector, intervening
state transitions, and target read vector. In mHC, the transition
matrices are doubly stochastic.

For Block AttnRes, we use $\bm k_i=\bm v_i$ and the learned
pseudo-query $\bm w_l$. As in AttnRes, the figure displays
unnormalized scores
$\phi(\bm w_l,\bm k)=
\exp(\bm w_l^\top\operatorname{RMSNorm}(\bm k))$.
The first completed block contributes the single source
$\bm k_1+\bm k_2$; its score spans both constituent columns.
Normalization counts this grouped source once, rather than once
per column. Dashed cells mark the evolving partial source.

Block MSAR retains this grouping and constructs one score matrix
per stream, using $\bm w_l^{(r)}$ and $\mathcal E_r[\bm b]$ in
place of $\bm w_l$ and $\bm b$.
Each stream is normalized independently:
\begin{equation}
    \alpha_l^{(r)}(\bm b)
    =
    \frac{\phi(\bm w_l^{(r)},\mathcal E_r[\bm b])}
    {\sum_{\bm b'\in\mathcal S_l}
    \phi(\bm w_l^{(r)},\mathcal E_r[\bm b'])}.
    \label{eq:discussion-mars-block-weight}
\end{equation}
The expanded sources also serve as values, and the retrieved streams
are combined using $g_l^{(r)}$.
The notation $\mathcal E_r$ suppresses the convolution-bank index,
whose assignment is defined in Appendix~\ref{app:parameter-overhead}.
The MSAR panel thus shows a family of per-stream score matrices,
not a scalar mixing matrix acting on unexpanded sources.

\end{CJK*}
\end{document}